\documentclass[runningheads]{llncs}
\usepackage{graphicx}
\usepackage{siunitx}
\usepackage{color}
\usepackage{amssymb}
\usepackage{makecell}
\usepackage{multirow}
\usepackage{lipsum}
\usepackage{caption}
\usepackage{subcaption}

\usepackage{pifont}
\newcommand{\bfsection}[1]{\noindent\textbf{#1.}}

\definecolor{MyBlue}{rgb}{0,0.08,0.5}
\definecolor{MyRed}{rgb}{0.7,0.02,0.02}
\definecolor{MyOrange}{rgb}{1,0.5,0}
\definecolor{MyPurple}{rgb}{0.6,0.25,0.8}
\definecolor{MyGreen}{rgb}{0.1,0.8,0.1}

\newcommand{\etal}{\textit{et al}.}
\newcommand{\ie}{\textit{i}.\textit{e}.}
\newcommand{\eg}{\textit{e}.\textit{g}.}

\usepackage[pagebackref=false,breaklinks=true,letterpaper=true,colorlinks,bookmarks=false]{hyperref}
\hypersetup{
     colorlinks = true,
     linkcolor = red,
     anchorcolor = black,
     citecolor = black,
     filecolor = black,
     urlcolor = black
     }

\newcommand\blfootnote[1]{%
  \begingroup
  \renewcommand\thefootnote{}\footnote{#1}%
  \addtocounter{footnote}{-1}%
  \endgroup
}

\begin{document}
\title{AxonEM Dataset: 3D Axon Instance Segmentation of Brain Cortical Regions}
\titlerunning{AxonEM Dataset}

\newcommand*{\affaddr}[1]{#1} 
\newcommand*{\affmark}[1][*]{\textsuperscript{#1}}

\author{Donglai Wei\affmark[1]$^{\dagger}$ \and Kisuk Lee\affmark[2]$^{\dagger}$ Hanyu Li\affmark[3] \and Ran Lu\affmark[2] \and J. Alexander Bae\affmark[2] \and Zequan Liu\affmark[4]$^{*}$ \and Lifu Zhang\affmark[5]$^{*}$ \and Márcia dos Santos\affmark[6]$^{*}$ \and Zudi Lin\affmark[1] \and Thomas Uram\affmark[7] \and Xueying Wang\affmark[1] \and Ignacio Arganda-Carreras\affmark[8,9,10] \and Brian Matejek\affmark[1] \and Narayanan Kasthuri\affmark[3,7] \and Jeff Lichtman\affmark[1] \and Hanspeter Pfister\affmark[1]}

\authorrunning{D. Wei and K. Lee {\em et al.}}
\institute{
\affaddr{\affmark[1] Harvard University} \
\affaddr{\affmark[2] Princeton University} \
\affaddr{\affmark[3] University of Chicago} \
\affaddr{\affmark[4] RWTH Aachen University} \
\affaddr{\affmark[5] Boston University} \
\affaddr{\affmark[6] Universidade do Vale do Rio dos Sinos}\
\affaddr{\affmark[7] Argonne National Laboratory} \
\affaddr{\affmark[8] Donostia International Physics Center} \
\affaddr{\affmark[9] University of the Basque Country (UPV/EHU)} \
\affaddr{\affmark[10] Ikerbasque, Basque Foundation for Science} \\
\email{donglai@seas.harvard.edu}
}

\maketitle   
\begin{abstract}
    Electron microscopy (EM) enables the reconstruction of neural circuits at the level of individual synapses, which has been transformative for scientific discoveries. 
    However, due to the complex morphology, an accurate reconstruction of cortical axons has become a major challenge. 
    Worse still, there is no publicly available large-scale EM dataset from the cortex that provides dense ground truth segmentation for axons, making it difficult to develop and evaluate large-scale axon reconstruction methods.
    To address this, we introduce the \textsl{AxonEM} dataset, which consists of two $30\times30\times30~\mu$m$^3$ EM image volumes from the human and mouse cortex, respectively. We thoroughly proofread over 18,000 axon instances to provide dense 3D axon instance segmentation, enabling large-scale evaluation of axon reconstruction methods. In addition, we densely annotate nine ground truth subvolumes for training, per each data volume. With this, we reproduce two published state-of-the-art methods and provide their evaluation results as a baseline.
    We publicly release our code and data at \url{https://connectomics-bazaar.github.io/proj/AxonEM/index.html} to foster the development of advanced methods. 
\end{abstract}

\blfootnote{$^{\dagger}$ Equal contribution.}
\blfootnote{$^{*}$ Works were done during the internship at Harvard University.}
\vspace{-0.4in}
\section{Introduction}
With recent technical advances in high-throughput 3D electron microscopy (EM)~\cite{kornfeld2018progress}, it has become feasible to map out neural circuits at the level of individual synapses in large brain volumes~\cite{dorkenwald2017automated,scheffer2020connectome,shapson2021connectomic}. Reconstruction of a wiring diagram of the whole-brain, or a connectome, has seen successes in invertebrates, such as the complete connectome of the roundworm \textit{C. elegans}~\cite{whiteCElegans} and the partial connectome of the fruit fly \textit{D. melanogaster}~\cite{dorkenwald2020flywire,scheffer2020connectome}. Mapping the connectome of a whole mouse brain is currently being considered as the next transformative challenge in the field of connectomics~\cite{abbott2020mind}. Detailed study of large-scale connectomes may lead to new scientific discoveries with potential clinical implications~\cite{abbott2020mind}.

\begin{figure}[t]
     \centering
     \includegraphics[width=1.0\textwidth]{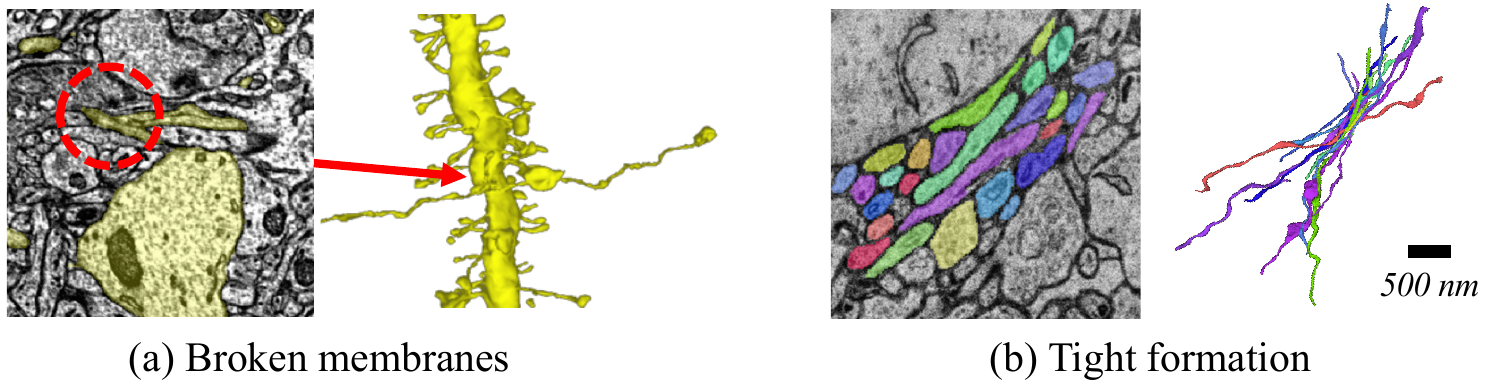}
     \caption{Example axons in our AxonEM dataset. \textbf{(a)} Due to the unclear cell boundaries resulted from imperfect staining, an axon is falsely merged with an abutting dendrite. \textbf{(b)} Axons often form a tight bundle. Due to the similar shape and small size of their cross-sections, axons in such a tight bundle may be prone to errors, especially in the presence of image misalignment.}
     \label{fig:teaser}
     \vspace{-0.18in}
\end{figure}

At present, automated reconstruction of neural circuits from the cortex faces a significant obstacle: the axon instance segmentation. In the cortex, axons are the most abundant type of neurites by path length~\cite{motta2019dense}, with the smallest diameter of $\sim$50~nm in the case of mouse cortical axons~\cite{helmstaedter2013cellular}. Given their small caliber, complex morphology, and densely packed nature (Fig.~\ref{fig:teaser}), axons are prone to reconstruction errors especially when challenged by various image defects such as imperfect staining and image misalignment~\cite{lee2019convolutional}.

To the best of our knowledge, there is no existing EM reconstruction from any cortical region that provides \textit{densely and fully} proofread axon instances. Previously the largest work was carried out by Motta \etal~\cite{motta2019dense}, in which the team spent $\sim$4,000 human hours in semi-automated proofreading of the initial automated segmentation. However, their finalized reconstruction for axons suffered from a relatively high error rate~\cite{motta2019dense},\footnote{12.8 errors per 1 mm of path length, estimated with 10 randomly chosen axons~\cite{motta2019dense}.} making their reconstruction less suitable for developing and evaluating large-scale axon reconstruction methods.

\begin{figure}[t]
     \centering
     \includegraphics[width=\textwidth]{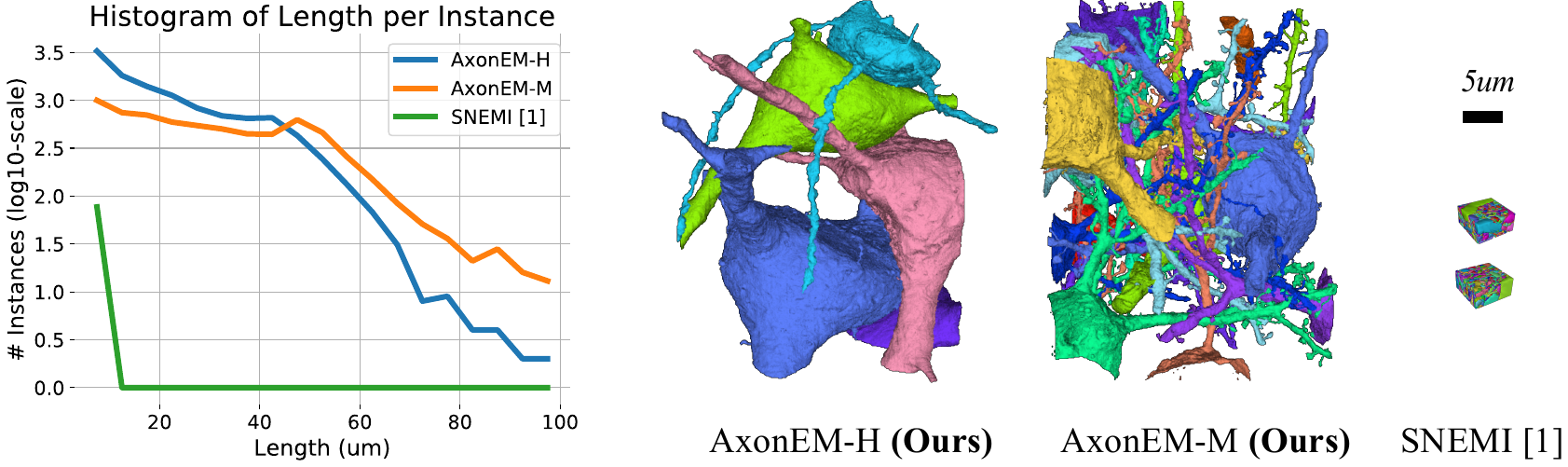}
    \caption{Comparison of EM datasets with proofread axon instances. (Left) Distribution of axon instance length. (Right) 3D rendering of neurons with somas in AxonEM dataset and the dense segmentation of SNEMI3D dataset~\cite{SNEMI3D}. Two large glial cells (not rendered) in AxonEM-H occupy the space between the blue and pink neurons, leading to much fewer long axons compared to AxonEM-M.}
    \label{fig:dataset_size}
    \vspace{-0.1in}
\end{figure}

To foster new research to address the challenge, we have created a large-scale benchmark for 3D axon instance segmentation, \textsl{AxonEM}, which is about 250$\times$ larger than the prior art~\cite{SNEMI3D} (Fig.~\ref{fig:dataset_size}). AxonEM consists of two $30 \times 30 \times 30~\mu$m$^3$ EM image volumes acquired from the mouse and human cortex, respectively. Both image volumes span, at least partially, layer 2 of the cortex, thus allowing for comparative studies on brain structures across different mammalian species. 

\bfsection{Contributions} First, we thoroughly proofread over 18,000 axon instances to provide the \textit{largest ever} ground truth for cortical axons, which enables large-scale \textit{evaluation}. Second, we \textit{densely} annotate a handful of dataset subvolumes, which can be used to \textit{train} a broad class of methods for axon reconstruction and allow for an objective comparison. Third, we publicly release our data and baseline code to reproduce two published state-of-the-art neuron instance segmentation methods~\cite{Januszewski2018FFN,lee2017superhuman}, thereby calling for innovative methods to tackle the challenge.
\subsection{Related Works}
\bfsection{Axon Segmentation}
In non-cortical regions, myelinated axons are thick and easier to segment. 
Traditional image processing algorithms have been proposed, including thresholding and morphological operations~\cite{cuisenaire1999automatic}, axon shape-based morphological discrimination~\cite{more2011semi}, watershed~\cite{wang2012segmentation}, region growing~\cite{zhao2010automatic}, active contours without~\cite{begin2014automated} and with discriminant analysis~\cite{zaimi2016axonseg}. 
More recently, deep learning methods have also been used to segment this type of axons
~\cite{naito2017identification,mesbah2016deep,zaimi2018axondeepseg}.

However, {\it unmyelinated} axons in the cortex have been a significant source of reconstruction errors, because their thin and intricate branches are vulnerable to various types of image defects~\cite{lee2019convolutional}. Motta \etal~\cite{motta2019dense} took a semi-automated reconstruction approach to densely segment axons in a volume of $\sim$500,000 $\mu$m$^3$ from the mouse cortex. However, their estimate of remaining axon reconstruction errors was relatively high~\cite{motta2019dense}.


\bfsection{Neuron Segmentation in EM}
There are two mainstream approaches to instance segmentation from 3D EM image volumes.
One approach trains convolutional neural networks (CNNs) to predict an intermediate representation for neuronal boundaries~\cite{ciresan2012deep,ronneberger2015u,zeng2017deepem3d} or inter-voxel affinities~\cite{turaga2010convolutional,lee2017superhuman,funke2018large},
which are then often conservatively oversegmented with watershed~\cite{zlateski2015image}. For further improvement, adjacent segments are agglomerated by a similarity measure~\cite{nunez2014graph,lee2017superhuman,funke2018large}, or with graph partitioning~\cite{beier2017multicut}. 
In the other approach, CNNs are trained to recursively extend a segmentation mask for a single object of interest~\cite{meirovitch2016multipass,Januszewski2018FFN}. Recently, recurrent neural networks (RNNs) have also been employed to extend multiple objects simultaneously~\cite{meirovitch2019cross,gonda2021consistent}.



\section{AxonEM Dataset}
The proposed AxonEM dataset contains two EM image volumes cropped from the published cortex datasets, one from mouse~\cite{dorkenwald2019binary} (AxonEM-M) and the other from human~\cite{shapson2021connectomic,wei2020mitoem} (AxonEM-H). We densely annotated axons longer than $5~\mu$m in both volumes (Sec.~\ref{sec:annotation}), and performed basic analysis (Sec.~\ref{sec:analysis}).

\subsection{Dataset Description}
Two tissue blocks were imaged with serial section EM from layer 2/3 in the primary visual cortex of an adult mouse and from layer 2 in the temporal lobe of an adult human. The image volumes were 2$\times$ in-plane-downsampled to the voxel resolutions of $7\times7\times40$~nm$^3$ and $8\times8\times30$~nm$^3$, respectively. We cropped out two $30 \times 30 \times 30~\mu$m$^3$ subvolumes, AxonEM-M and AxonEM-H, intentionally avoiding large blood vessels in order to contain more axon instances. We refer readers to the original papers for more dataset details~\cite{dorkenwald2019binary,shapson2021connectomic}.
\begin{figure}[t]
     \centering
     \includegraphics[width=\textwidth]{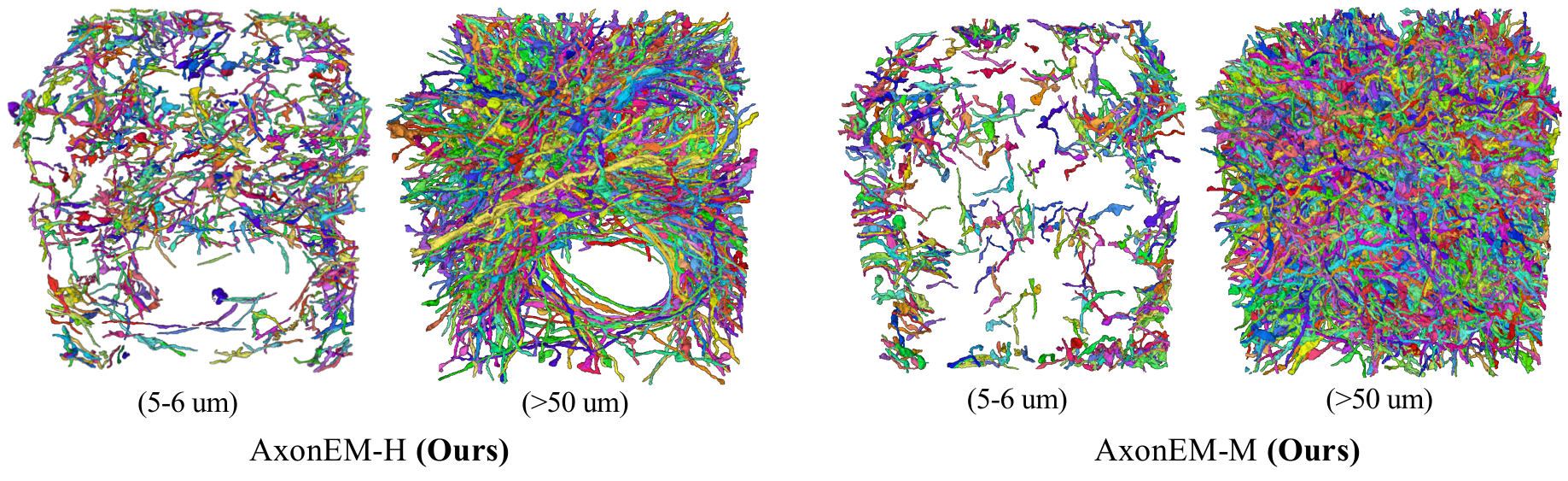}\\
     \caption{Visualization of ground truth axon segmentation. We show 3D meshes of axons with length either 5--6$~\mu$m or $>50~\mu$m. As expected, the short axons are fragments near the volume boundaries that briefly enter and leave the volume.}
     \label{fig:dataset_mesh}
\end{figure}

\subsection{Dataset Annotation}
\label{sec:annotation}
For training, we annotated nine $512\times512\times50$ subvolumes to produce dense ground truth segmentation, respectively, for each volume.
For evaluation, we annotated all axon segments that are longer than 5~$\mu$m. We choose this threshold because segments shorter than 5~$\mu$m mostly concentrate near the borders of the volumes, making it difficult to confidently classify them into axons due to the lack of context. We took a semi-automatic approach for annotation. Starting from the initial segmentation provided by the original datasets~\cite{dorkenwald2019binary,shapson2021connectomic}, proofreaders inspected potential error locations suggested by automatic detection~\cite{matejek2019biologically} in Neuroglancer~\cite{ng}, and manually corrected remaining errors in VAST~\cite{berger2018vast}.

\bfsection{Initial Dense Segmentation}
For AxonEM-M, we downloaded a publicly available volume segmentation computed from a variant of 3D U-Net model~\cite{lee2017superhuman,dorkenwald2019binary} and applied the connected components method to obtain the initial segmentation to start with.
For AxonEM-H, we manually created small volumes of dense segmentation to train the flood-filling network~\cite{Januszewski2018FFN} and ran inference on the entire AxonEM-H volume to generate the initial segmentation to start with.


\bfsection{Axon Identification}
We first identified axons from the initial dense segmentation, along with the skeletonization obtained by the TEASER algorithm~\cite{TEASAR}. We computed for each segment the following statistics: volume, path length, and width (\ie,~average radius of all skeleton points). Based on these stats, we used the K-means method to cluster the segments into $K=10$ groups with similar morphology. Then we visually inspected the 3D meshes of all segments that are longer than 5~$\mu$m, and classified them into four categories: glia, dendrite, axon, or uncertain. Instead of going through each individual segments one by one, we pulled a small-sized batch from each of the groups with similar morphology. This enabled us to collectively inspect such a ``minibatch'' of morphologically similar segments all at once, thus making our classification more efficient and effective.

\bfsection{Axon Proofreading}
After identifying axons, we automatically detected the potential error locations based on morphological features and biological prior knowledges~\cite{matejek2019biologically}. To correct remaining split errors, we measured the direction of the axon skeletons at every end point, and then selected candidate pairs of axons whose end points are close in space and pointing to each other in direction. We examined these candidate pairs, as well as those axon skeletons that terminate in the middle of the volume without reaching the boundaries, which may indicate a premature termination caused by a split error. To fix remaining merge errors, we inspected the junction points in the skeleton graph, because some of them may exhibit a biologically implausible motif such as the ``X-shaped'' junction~\cite{meirovitch2016multipass}.


\bfsection{Finalization}
Four annotation experts with experience in axon reconstruction were recruited to proofread and cross-check the axon identification and proofreading results until there remains no disagreement between the annotators.

\subsection{Dataset Analysis}
\label{sec:analysis}
The physical size of our EM volumes is about 250$\times$ larger than the previous SNEMI3D benchmark~\cite{SNEMI3D}.
AxonEM-H and AxonEM-M have around 11.3K and 6.7K axon instances (longer than 5~$\mu$m), respectively, over 200$\times$ more than that of SNEMI3D.
We compare the distribution of axon instance length in Fig.~\ref{fig:dataset_size}, where our AxonEM dataset contains more and longer axons. Due to the presence of two glial cells, in AxonEM-H there are fewer long, branching axons than in AxonEM-M.
In Fig.~\ref{fig:dataset_mesh}, we show 3D meshes of short (5--6~$\mu$m) and long axons ($>$50 $\mu$m) from both volumes. 
Again, one can see the vacancy of long, branching axons in AxonEM-H where the glial cells (not visualized) occupy the space.
\section{Methods}

For 3D axon instance segmentation, we describe our choice of evaluation metric (Sec.~\ref{sec:task}) and two state-of-the-art methods adopted for later benchmark (Sec.~\ref{sec:exp}).


\subsection{Task and Evaluation Metric}\label{sec:task}
Our task is dense segmentation of 3D EM images of the brain, \eg, as in the SNEMI3D benchmark challenge~\cite{SNEMI3D}. We focus, however, our attention only to the accurate reconstruction of axons, which are challenging due to their long, thin morphology and complicated branching patterns. To this end, our evaluation is restricted to axons among all ground truth segments.



To measure the overall accuracy of axon reconstruction, we adopted the expected run length (ERL,~\cite{Januszewski2018FFN}) to estimate the average error-free path length of the reconstructed axons. 
The previous ARAND metric used in SNEMI3D~\cite{SNEMI3D} requires dense ground truth segmentation with voxel-level precision, which is impractical for large-scale datasets.
In contrast, the original ERL metric focuses on the skeleton-level accuracy with a disproportionately larger penalty on merge errors, in which two or more distinct objects are merged together erroneously.
In practice, due to skeletonization artifacts, the original ERL metric may assign a good axon segment \emph{zero} run length if there exists a merge error with just one outlier skeleton point that is mistakenly placed near the ground truth segment boundaries. To mitigate this, we extended the original ERL metric, which is not robust to outlier skeleton nodes, to have a ``tolerance'' threshold of 50 skeleton nodes (around 2 $\mu$m in length) to relax the condition to determine whether the ground truth skeleton encounters a merge error in a segmentation.
\subsection{State-of-the-Art Methods}
We consider the top two methods for neuron instance segmentation on the SNEMI3D benchmark~\cite{SNEMI3D}: U-Net models predicting the affinity graph (affinity U-Net\footnote{\url{https://github.com/seung-lab/DeepEM}}~\cite{lee2017superhuman}) and flood-filling networks (FFN\footnote{\url{https://github.com/google/ffn}}~\cite{Januszewski2018FFN}). 
For model details, we refer readers to the original papers. Note that although their implementations are publicly available, in practice, a substantial amount of engineering effort may still be required in order to reproduce the reported performance.



\section{Experiments}\label{sec:exp}
\subsection{Implementation Details}\label{exp:details}
%

For the FFN pipeline, we used the default configuration in~\cite{Januszewski2018FFN} (convolution module depth=9, FOV size=[33, 33, 17], movement deltas=[8, 8, 4]) and trained separate models from scratch until accuracy saturates on the training data (94\% for AxonEM-M, 97\% for AxonEM-H, and 92\% for SNEMI3D). We then performed a single pass of inference without the over-segmentation consensus and FFN agglomeration steps~\cite{Januszewski2018FFN}. For inference, we ran distributed FFN on $512 \times 512 \times256$ voxel subvolumes with $64 \times 64 \times 32$ voxel overlap and later reconciliated into one volume~\cite{vescovi2020toward}. Training and inference were conducted on a multi-GPU cluster and reconciliation was done on a workstation \cite{vescovi2020toward}. Overall, for AxonEM-M and AxonEM-H, we used 32 NVidia A100 GPUs, spent 20 hours on training and 6.2 and 5.8 hours on inference for each dataset. For SNEMI3D, we trained the net for 10 hours with 16 NVidia A100 GPUs, then finished inference in 0.28 hours. 

For the affinity U-Net pipeline, we first used the Google Cloud Platform (GCP) compute engine to predict the affinity, using 12 NVidia T4 GPUs, each paired with a n1-highmem-8 instance (8 vCPUs and 52 GB memory). Processing the AxonEM-M and AxonEM-H datasets took 2.6 and 3.5 hours, respectively. From the predicted affinities, we produced segmentation using the watershed-type algorithm~\cite{zlateski2015image} and mean affinity agglomeration~\cite{lee2017superhuman}. We used 8 e2-standard-32 instances (32 vCPUs and 128 GB memory). The watershed and agglomeration steps together took 0.58 hours for AxonEM-M, and 0.75 hours for AxonEM-H.


\begin{table}[t]
\caption{\label{tab:snemi3d}Segmentation results on SNEMI3D dataset. We report the ARAND score for the dense segmentation and ERL score for axon segments among them. The percentage numbers in parentheses represent the ratio of the baseline axon ERL with respect to the upper limit given by the ground truth ERL.} 
\centering
\begin{tabular}{llccc}
\hline
\multicolumn{2}{l}{\bf Method} & {\bf~ARAND$\downarrow$~} &~ {\bf ERL$\uparrow$ ($\mu$m)}\\
    \hline
    \hline
    \multicolumn{2}{l}{Ground truth} & 0.000 & 8.84 (100\%)\\
    \hline
    \hline
    \multicolumn{2}{l}{PNI's Aff. U-Net (mean aff. aggl.)~\cite{lee2017superhuman}} & 0.031  & 8.31 (94\%)\\
    \multicolumn{2}{l}{Google's FFN~\cite{Januszewski2018FFN}} & 0.029  &  8.20 (93\%)\\
    \hline
    \hline
    \multirow{2}{*}{\bf Our Impl.} & ~~~Aff. U-Net (mean aff. aggl.) & 0.038  & 8.22 (93\%) \\
    & ~~~FFN (w/o aggl.)  & 0.112 & 5.18 (59\%) \\
    \hline
\end{tabular}
\end{table}

\label{exp:lucchi}
\subsection{Benchmark Results on SNEMI3D Dataset}

We first show the results of our reproducing experiments on the SNEMI3D benchmark in Table~\ref{tab:snemi3d}, where we obtained the adapted Rand (ARAND) error scores from the challenge evaluation server. Our affinity U-Net error score (0.038) was comparable to that of \cite{lee2017superhuman} (0.031), confirming that we are capable of faithfully reproducing the affinity U-Net method.
However, our FFN error score (0.112) was severely worse than that of \cite{Januszewski2018FFN} (0.033) by a large margin, suggesting that one should not consider our FFN baseline as state-of-the-art.

Next, we computed the axon ERL for our reproducing results, as well as for the ground truth segmentation provided by the challenge organizer and some of the top submission results \cite{lee2017superhuman,Januszewski2018FFN}. The ERL for the ground truth axons was 8.84 $\mu$m, which is the average path length of the ground truth axons. As expected from the worst ARAND error score, our FFN result showed the lowest ERL (5.18 $\mu$m) among others. The axon ERLs for the top submission results \cite{lee2017superhuman,Januszewski2018FFN} and our affinity U-Net result were all above 8 $\mu$m, approaching the upper limit given by the ground truth axon ERL (8.84 $\mu$m). Interestingly, a lower (better) ARAND error score for dense segmentation does not necessarily result in a higher (better) axon ERL. This is because the ARAND error score is computed for dense segmentation including all segments (axon, dendrite, and glia), whereas we computed the ERL selectively for axon. For example, our affinity U-Net result was better in axon ERL, but worse in ARAND, than \cite{Januszewski2018FFN} because of more problems in dendrite and glia.
\begin{table}[t]
\caption{\label{tab:exp_main}Axon segmentation results on AxonEM dataset (ERL).} 
\centering
\begin{tabular}{llccc}
\hline
\multicolumn{2}{l}{\bf Method} & {\bf AxonEM-H} &~ {\bf AxonEM-M}\\
    \hline
    \hline
    \multicolumn{2}{l}{Ground truth} & 28.5 $\mu$m (100\%) & 43.6 $\mu$m (100\%)\\
    \hline
    \hline
    \multirow{2}{*}{\bf Our Impl.} & ~~~Aff. U-Net (mean aff. aggl.) & 18.5 $\mu$m (65\%)  & 38.5 $\mu$m (88\%) \\
    &~~~FFN (w/o aggl.) & 9.6 $\mu$m (34\%) &  9.4 $\mu$m (22\%)\\
    \hline
\end{tabular}
\end{table}

\begin{figure}[t]
     \centering
     \includegraphics[width=\textwidth]{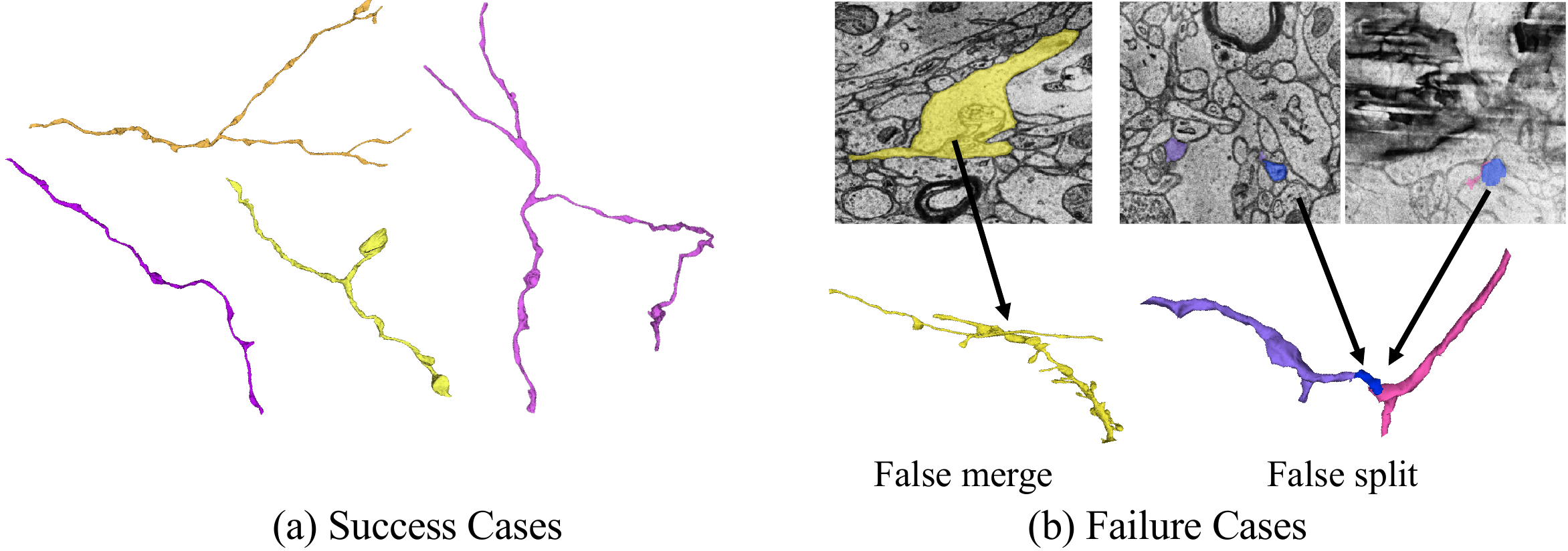}
     \caption{Qualitative results of our affinity U-Net baseline on AxonEM. Failure cases are often caused by invaginating structures, thin parts, and image defects.}
     \label{fig:exp_qual}
\end{figure}
\subsection{Benchmark Results on AxonEM Dataset}
\label{exp:main}
We provide axon segmentation baselines for the AxonEM dataset by reproducing two state-of-the-art methods, affinity U-Net \cite{lee2017superhuman} and FFN \cite{Januszewski2018FFN}. One should be cautious that our FFN baseline does not fully reproduce the state-of-the-art accuracy of the original method by \cite{Januszewski2018FFN}, due to the reasons detailed below.

\bfsection{Quantitative Results}
As shown in Table~\ref{tab:exp_main}, our affinity U-Net baseline achieved a relatively high axon ERL (88\% with respect to the ground truth) on AxonEM-M, whereas its ERL for AxonEM-H was relatively lower (65\%). The axon ERL of our FFN baseline was consistently and significantly lower (22\% on AxonEM-H and 34\% on AxonEM-H) than our affinity U-Net baseline, being consistent with our SNEMI3D reproducing results in the previous section (Table~\ref{tab:snemi3d}). 
Due to the lack of open-source code for several inference steps, our FFN baseline was not a faithful reproduction of the full reconstruction pipeline proposed by \cite{Januszewski2018FFN}. 
Our FFN baseline lacks both the multi-scale over-segmentation consensus and FFN agglomeration steps in \cite{Januszewski2018FFN}, which are crucial in achieving a very low merge error rate while maximizing ERL. A much better FFN baseline by faithfully reproducing the full reconstruction pipeline remains to be done in the future.

\bfsection{Qualitative Results}
We show both successful and unsuccessful axon segmentation results of our affinity U-Net baselines (Fig.~\ref{fig:exp_qual}).
Due to the thin parts, the axon reconstruction is more vulnerable to image defects like missing section, staining artifact, knife mark, and misalignment~\cite{lee2019convolutional} than other structures.


\section{Conclusion}
In this paper, we have introduced a large-scale 3D EM dataset for axon instance segmentation, and provided baseline results by re-implementing two state-of-the-art methods. 
We expect that the proposed densely annotated large-scale axon dataset will not only foster development of better axon segmentation methods, but also serve as a test bed for various other applications.

\subsubsection*{Acknowledgment.}
KL, RL, and JAB were supported by the Intelligence Advanced Research Projects Activity (IARPA) via Department of Interior/ Interior Business Center (DoI/IBC) contract number D16PC0005, NIH/NIMH (U01MH114824, U01MH117072, RF1 MH117815), NIH/NINDS (U19NS104648, R01NS104926), NIH/NEI (R01EY027 036), and ARO (W911NF-12-1-0594), and are also grateful for assistance from Google, Amazon, and Intel.
DW, ZL, JL, and HP were partially supported by NSF award IIS-1835231.
I. A-C would like to acknowledge the support of the Beca Leonardo a Investigadores y Creadores Culturales 2020 de la Fundación BBVA.
We thank Viren Jain, Michał Januszewski and their team for generating the initial segmentation for AxonEM-H, and Daniel Franco-Barranco for setting up the challenge using AxonEM.

\bfsection{Declaration of Interests} KL, RL, and JAB disclose financial interests in Zetta AI LLC.
\bibliographystyle{splncs04}
\bibliography{egbib}

\begin{thebibliography}{10}
\providecommand{\url}[1]{\texttt{#1}}
\providecommand{\urlprefix}{URL }
\providecommand{\doi}[1]{https://doi.org/#1}

\bibitem{ng}
Neuroglancer, \url{https://github.com/google/neuroglancer}

\bibitem{SNEMI3D}
{SNEMI3D EM} segmentation challenge and dataset,
  \url{http://brainiac2.mit.edu/SNEMI3D/home}

\bibitem{abbott2020mind}
Abbott, L.F., Bock, D.D., Callaway, E.M., Denk, W., Dulac, C., Fairhall, A.L.,
  Fiete, I., Harris, K.M., et~al.: The mind of a mouse. Cell  (2020)

\bibitem{begin2014automated}
B{\'e}gin, S., Dupont-Therrien, O., B{\'e}langer, E., Daradich, A., Laffray,
  S., et~al.: Automated method for the segmentation and morphometry of nerve
  fibers in large-scale cars images of spinal cord tissue. Biomedical optics
  express  (2014)

\bibitem{beier2017multicut}
Beier, T., Pape, C., Rahaman, N., Prange, T., Berg, S., Bock, D.D., Cardona,
  A., Knott, G.W., Plaza, S.M., Scheffer, L.K., et~al.: Multicut brings
  automated neurite segmentation closer to human performance. Nature methods
  (2017)

\bibitem{berger2018vast}
Berger, D.R., Seung, H.S., Lichtman, J.W.: Vast (volume annotation and
  segmentation tool): efficient manual and semi-automatic labeling of large
  3{D} image stacks. Frontiers in neural circuits  (2018)

\bibitem{ciresan2012deep}
Ciresan, D., Giusti, A., Gambardella, L.M., Schmidhuber, J.: Deep neural
  networks segment neuronal membranes in electron microscopy images. In:
  NeurIPS (2012)

\bibitem{cuisenaire1999automatic}
Cuisenaire, O., Romero, E., Veraart, C., Macq, B.M.: Automatic segmentation and
  measurement of axons in microscopic images. In: Medical Imaging (1999)

\bibitem{dorkenwald2020flywire}
Dorkenwald, S., McKellar, C., et~al.: Flywire: Online community for whole-brain
  connectomics. bioRxiv  (2020)

\bibitem{dorkenwald2017automated}
Dorkenwald, S., Schubert, P.J., Killinger, M.F., Urban, G., Mikula, S., Svara,
  F., Kornfeld, J.: Automated synaptic connectivity inference for volume
  electron microscopy. Nature methods  (2017)

\bibitem{dorkenwald2019binary}
Dorkenwald, S., Turner, N.L., Macrina, T., Lee, K., Lu, R., Wu, J., Bodor,
  A.L., Bleckert, A.A., Brittain, D., Kemnitz, N., et~al.: Binary and analog
  variation of synapses between cortical pyramidal neurons. BioRxiv  (2019)

\bibitem{funke2018large}
Funke, J., Tschopp, F., Grisaitis, W., Sheridan, A., Singh, C., Saalfeld, S.,
  Turaga, S.C.: Large scale image segmentation with structured loss based deep
  learning for connectome reconstruction. TPAMI  (2018)

\bibitem{gonda2021consistent}
Gonda, F., Wei, D., Pfister, H.: Consistent recurrent neural networks for 3d
  neuron segmentation. In: ISBI (2021)

\bibitem{helmstaedter2013cellular}
Helmstaedter, M.: Cellular-resolution connectomics: challenges of dense neural
  circuit reconstruction. Nature Methods  (2013)

\bibitem{Januszewski2018FFN}
Januszewski, M., Kornfeld, J., Li, P.H., Pope, A., Blakely, T., Lindsey, L.,
  Maitin-Shepard, J., Tyka, M., Denk, W., Jain, V.: High-precision automated
  reconstruction of neurons with flood-filling networks. Nature Methods  (2018)

\bibitem{kornfeld2018progress}
Kornfeld, J., Denk, W.: Progress and remaining challenges in high-throughput
  volume electron microscopy. Current Opinion in Neurobiology  (2018)

\bibitem{lee2019convolutional}
Lee, K., Turner, N., Macrina, T., Wu, J., Lu, R., Seung, H.S.: Convolutional
  nets for reconstructing neural circuits from brain images acquired by serial
  section electron microscopy. Current Opinion in Neurobiology  (2019)

\bibitem{lee2017superhuman}
Lee, K., Zung, J., Li, P., Jain, V., Seung, H.S.: Superhuman accuracy on the
  snemi3d connectomics challenge. arXiv:1706.00120  (2017)

\bibitem{matejek2019biologically}
Matejek, B., Haehn, D., Zhu, H., Wei, D., Parag, T., Pfister, H.:
  Biologically-constrained graphs for global connectomics reconstruction. In:
  CVPR (2019)

\bibitem{meirovitch2019cross}
Meirovitch, Y., Mi, L., Saribekyan, H., Matveev, A., Rolnick, D., Shavit, N.:
  Cross-classification clustering: An efficient multi-object tracking technique
  for 3-d instance segmentation in connectomics. In: CVPR (2019)

\bibitem{meirovitch2016multipass}
Meirovitch, Y., et~al.: A multi-pass approach to large-scale connectomics.
  arXiv preprint arXiv:1612.02120  (2016)

\bibitem{mesbah2016deep}
Mesbah, R., McCane, B., Mills, S.: Deep convolutional encoder-decoder for
  myelin and axon segmentation. In: IVCNZ (2016)

\bibitem{more2011semi}
More, H.L., Chen, J., Gibson, E., Donelan, J.M., Beg, M.F.: A semi-automated
  method for identifying and measuring myelinated nerve fibers in scanning
  electron microscope images. Journal of neuroscience methods  (2011)

\bibitem{motta2019dense}
Motta, A., Berning, M., Boergens, K.M., Staffler, B., Beining, M., Loomba, S.,
  Hennig, P., Wissler, H., Helmstaedter, M.: Dense connectomic reconstruction
  in layer 4 of the somatosensory cortex. Science  (2019)

\bibitem{naito2017identification}
Naito, T., Nagashima, Y., Taira, K., Uchio, N., Tsuji, S., Shimizu, J.:
  Identification and segmentation of myelinated nerve fibers in a
  cross-sectional optical microscopic image using a deep learning model.
  Journal of neuroscience methods  (2017)

\bibitem{nunez2014graph}
Nunez-Iglesias, J., Kennedy, R., Parag, T., Shi, J., Chklovskii, D.B.: Machine
  learning of hierarchical clustering to segment 2d and 3d images. PloS one
  (2013)

\bibitem{ronneberger2015u}
Ronneberger, O., Fischer, P., Brox, T.: U-net: Convolutional networks for
  biomedical image segmentation. In: MICCAI (2015)

\bibitem{TEASAR}
Sato, M., Bitter, I., Bender, M., Kaufman, A., Nakajima, M.: {TEASAR}:
  tree-structure extraction algorithm for accurate and robust skeletons. In:
  Pacific Conference on Computer Graphics and Applications

\bibitem{scheffer2020connectome}
Scheffer, L.K., Xu, C.S., Januszewski, M., Lu, Z., Takemura, S.y., Hayworth,
  K.J., Huang, G.B., Shinomiya, K., Maitlin-Shepard, J., Berg, S., et~al.: A
  connectome and analysis of the adult drosophila central brain. Elife  (2020)

\bibitem{shapson2021connectomic}
Shapson-Coe, A., et~al.: A connectomic study of a petascale fragment of human
  cerebral cortex. bioRxiv  (2021)

\bibitem{turaga2010convolutional}
Turaga, S.C., Murray, J.F., Jain, V., Roth, F., Helmstaedter, M., Briggman, K.,
  Denk, W., Seung, H.S.: Convolutional networks can learn to generate affinity
  graphs for image segmentation. Neural computation  (2010)

\bibitem{vescovi2020toward}
Vescovi, R., Li, H., Kinnison, J., Ke{\c{c}}eli, M., Salim, M., Kasthuri, N.,
  Uram, T.D., Ferrier, N.: Toward an automated hpc pipeline for processing
  large scale electron microscopy data. In: XLOOP (2020)

\bibitem{wang2012segmentation}
Wang, Y.Y., Sun, Y.N., Lin, C.C.K., Ju, M.S.: Segmentation of nerve fibers
  using multi-level gradient watershed and fuzzy systems. AI in Medicine
  (2012)

\bibitem{wei2020mitoem}
Wei, D., Lin, Z., Franco-Barranco, D., et~al.: Mito{EM} dataset: Large-scale 3d
  mitochondria instance segmentation from em images. In: MICCAI (2020)

\bibitem{whiteCElegans}
White, J.G., Southgate, E., Thomson, J.N., Brenner, S.: The structure of the
  nervous system of the nematode caenorhabditis elegans. Philosophical
  Transactions of the Royal Society B: Biological Sciences  (1986)

\bibitem{zaimi2016axonseg}
Zaimi, A., Duval, T., Gasecka, A., C{\^o}t{\'e}, D., Stikov, N., Cohen-Adad,
  J.: Axon{S}eg: open source software for axon and myelin segmentation and
  morphometric analysis. Frontiers in neuroinformatics  (2016)

\bibitem{zaimi2018axondeepseg}
Zaimi, A., Wabartha, M., Herman, V., Antonsanti, P.L., Perone, C.S.,
  Cohen-Adad, J.: Axondeepseg: automatic axon and myelin segmentation from
  microscopy data using convolutional neural networks. Scientific reports
  (2018)

\bibitem{zeng2017deepem3d}
Zeng, T., Wu, B., Ji, S.: Deepem3d: approaching human-level performance on 3d
  anisotropic em image segmentation. Bioinformatics  (2017)

\bibitem{zhao2010automatic}
Zhao, X., Pan, Z., Wu, J., Zhou, G., Zeng, Y.: Automatic identification and
  morphometry of optic nerve fibers in electron microscopy images. Computerized
  Medical Imaging and Graphics  (2010)

\bibitem{zlateski2015image}
Zlateski, A., Seung, H.S.: Image segmentation by size-dependent single linkage
  clustering of a watershed basin graph. arXiv:1505.00249  (2015)

\end{thebibliography}

\end{document}